\documentclass[10pt,twocolumn,letterpaper]{article}

\usepackage{cvpr}              

\PassOptionsToPackage{sort, numbers, compress}{natbib}
\usepackage[numbers]{natbib}

\usepackage[utf8]{inputenc} %
\usepackage[T1]{fontenc}    %
\usepackage{url}            %
\usepackage{booktabs}       %
\usepackage{amsfonts}       %
\usepackage{amssymb}
\usepackage{nicefrac}       %
\usepackage{xcolor}
\usepackage{comment}
\usepackage{graphicx} 
\usepackage{caption,subcaption}
\usepackage{overpic}
\usepackage{amsmath}
\usepackage{physics}
\usepackage{color}
\usepackage{wrapfig}
\usepackage{colortbl}
\usepackage{textcomp}

\usepackage{booktabs} %
\usepackage{multirow}
\usepackage{tabularx}
\usepackage{amsthm}
\usepackage{lipsum,capt-of}

\usepackage{soul}
\usepackage{color}
\usepackage{xcolor}
\definecolor{turquoise}{cmyk}{0.65,0,0.1,0.3}
\definecolor{purple}{rgb}{0.65,0,0.65}
\definecolor{dark_green}{rgb}{0, 0.5, 0}
\definecolor{orange}{rgb}{0.8, 0.6, 0.2}
\definecolor{red}{rgb}{0.8, 0.2, 0.2}
\definecolor{darkred}{rgb}{0.6, 0.1, 0.05}
\definecolor{blueish}{rgb}{0.0, 0.3, .6}
\definecolor{light_gray}{rgb}{0.7, 0.7, .7}
\definecolor{pink}{rgb}{1, 0, 1}
\definecolor{greyblue}{rgb}{0.25, 0.25, 1}
\definecolor{tab_blue}{HTML}{1f77b4}
\definecolor{tab_orange}{HTML}{ff7f0e}
\definecolor{LightRed}{rgb}{0.99,0.89,0.89}
\definecolor{mesh_misty_rose}{HTML}{e6aaa3}
\definecolor{mesh_yellow}{HTML}{ffba00}
\definecolor{MyDarkBlue}{rgb}{0.02,0.02,0.6}

\newcommand*{\methodname}{MARBLE}

\newcommand{\TODO}[1]{\textbf{\color{red}[TODO: #1]}}

\newcommand{\TC}[1]{{\color{MyDarkBlue}[TC: #1]}}
\newcommand{\PS}[1]{{\color{tab_orange}[PS: #1]}}
\newcommand{\MB}[1]{{\color{dark_green}[MB: #1]}}
\newcommand{\VJ}[1]{{\color{darkred}[VJ: #1]}}

\renewcommand{\TODO}[1]{}

\renewcommand{\TC}[1]{}
\renewcommand{\PS}[1]{}
\renewcommand{\MB}[1]{}
\renewcommand{\VJ}[1]{}

\definecolor{cvprblue}{rgb}{0.21,0.49,0.74}
\usepackage[pagebackref,breaklinks,colorlinks,allcolors=cvprblue]{hyperref}
\usepackage{amsmath}

\def\paperID{23} 
\def\confName{CVPR}
\def\confYear{2025}

\newcommand\blfootnote[1]{%
  \begingroup
  \renewcommand\thefootnote{}\footnote{#1}%
  \addtocounter{footnote}{-1}%
  \endgroup
}

\title{MARBLE: Material Recomposition and Blending in CLIP-Space
}

\author{Ta Ying Cheng*\\
University of Oxford\\
{\tt\small}
\and
Prafull Sharma\\
MIT CSAIL\\
{\tt\small}
\and 
Mark Boss \\
Stability AI
{\tt\small}
\and
Varun Jampani \\
Stability AI
{\tt\small}
}

\begin{document}



\twocolumn[{
    \renewcommand\twocolumn[1][]{#1}
    \maketitle
    \begin{center}
    \centering
    \captionsetup{type=figure}
    \vspace{-2em}
    \includegraphics[width=\linewidth]{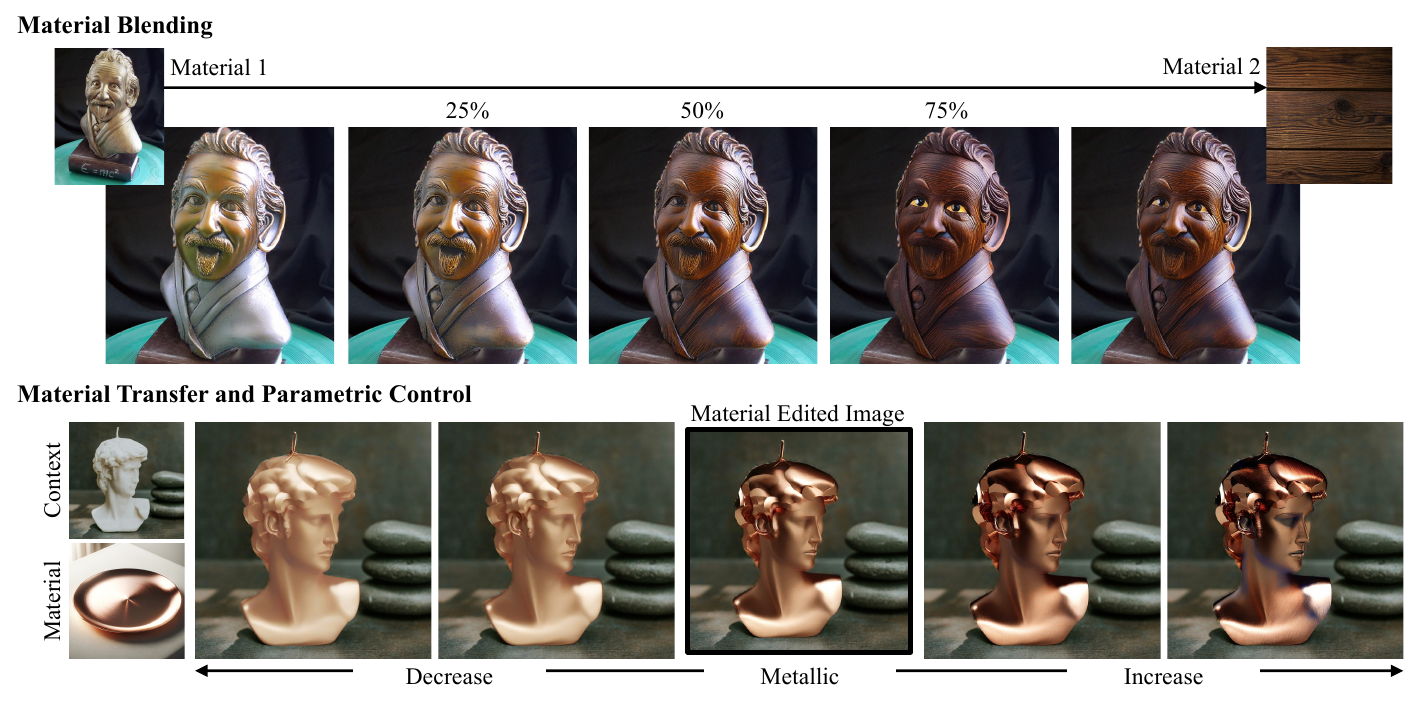}
    \vspace{-1em}
    \captionof{figure}{
        \textbf{Overview.} We present \methodname{}, a method for performing various material editing in images such as material blending (top row) and parametric control of material properties (bottom row) leveraging CLIP-space and pre-trained generative models. Given two material exemplar images, we can achieve a controllable blend of materials on the object by blending the material representation in CLIP-space. For parametric material attribute control, we learn a shallow network using synthetic data to predict the direction in CLIP-space for changing specific material properties such as metallic.
    }
    \label{fig:teaser}
    \end{center}
}]
\begin{abstract}
Editing materials of objects in images based on exemplar images is an active area of research in computer vision and graphics. We propose \methodname{}, a method for performing material blending and recomposing fine-grained material properties by finding material embeddings in CLIP-space and using that to control pre-trained text-to-image models. We improve exemplar-based material editing by finding a block in the denoising UNet responsible for material attribution. Given two material exemplar-images, we find directions in the CLIP-space for blending the materials. Further, we can achieve parametric control over fine-grained material attributes such as roughness, metallic, transparency, and glow using a shallow network to predict the direction for the desired material attribute change. We perform qualitative and quantitative analysis to demonstrate the efficacy of our proposed method. We also present the ability of our method to perform multiple edits in a single forward pass and applicability to painting.

\noindent Project Page: \href{https://marblecontrol.github.io/}{https://marblecontrol.github.io/}
\end{abstract} 
\blfootnote{\textsuperscript{*}Work was done during internship at Stability AI.}

\section{Introduction}~\label{sec:intro}












Editing object materials such as diffuse albedo, roughness, etc. in images is instrumental for graphics and vision applications such as game design, advertising, and visual content creation. Performing material editing in a single image using traditional graphics techniques requires understanding of several object and environment properties such as object geometry, its material properties as well as environment illumination, making it a highly challenging task. Previous approaches for material editing use crude approximations of object geometry and environment maps~\cite{khan2006image}, resulting in non-photorealistic results limited to finite material editing options. 

In this work, we tackle the problem of material transfer and recompose the material properties given a single image by directly leveraging the implicit knowledge of object and environment properties present in the pre-trained image diffusion models~\cite{rombach2021high}.
This circumvents the need for explicit estimation of these properties, which is challenging given a single image.
Recent works such as Alchemist~\cite{sharma2023alchemist} and ZeST~\cite{cheng2025zest} demonstrate the use of diffusion models for material editing in images.
ZeST~\cite{cheng2025zest} proposes a zero-shot technique for exemplar-based material transfer, where the object material from an exemplar image is transferred to the target object in the input image.
However, this approach is limited to high-level material changes and does not perform fine-grained control of material properties.
On the other hand, Alchemist~\cite{sharma2023alchemist} proposes a supervised fine-tuning of Stable Diffusion~\cite{rombach2021high} for fine-grained material control such as roughness, transparency etc. in images.
However, such a fine-tuning of the diffusion model has the potential to overfit to the synthetic data used for training, thereby destroying the valuable object prior knowledge in these models. 


In contrast, we propose a technique that can perform versatile material editing with material transfer as well as fine-grained control, while also retaining the base diffusion model priors.
In particular, we propose to keep the image diffusion model intact and perform material editing via CLIP~\cite{radford2021learning} image features that are injected into the diffusion model.
A key contribution of this work is to demonstrate that a surprising amount of material editing is possible with the manipulation of the CLIP-Space features.
Our technique called \methodname{} (Material Recomposition and Blending in CLIP-space) enables versatile material editing tasks ranging from 
performing coarse material transfer using an example image or blending materials from multiple objects (Figure \ref{fig:teaser} top row) to fine-grained material control of properties such as metallic, transparency, etc. (Figure \ref{fig:teaser} bottom row). 






It is far from trivial to achieve such diverse material editing tasks using only CLIP image features as CLIP captures all the object properties such as semantics, geometry etc., not just the material properties.
We build our method using ZeST architecture~\cite{cheng2025zest} for exemplar-based material transfer with some modifications. ZeST uses IP-Adapter~\cite{ye2023ip} that injects CLIP features into the diffusion model, along with a color-agnostic inpainting technique for material transfer from an exemplar image to the target object image.
With systematic experiments, we find a U-Net block in the Stable Diffusion that responds to the object materials. Following this insight, we propose to inject CLIP features into this specific U-Net block resulting in better material transfer. This modified architecture acts as the base for two variants of material editing. First, we show that this technique can also be used for material blending between two or more exemplar images.
For the fine-grained control of the material properties such as increasing or decreasing transparency, we propose to learn lightweight MLPs, using a small synthetic dataset, that predicts material editing directions for each of the individual material properties in the CLIP-space. As a result, we can achieve fine-grained control of the material properties by moving the CLIP features along these editing directions. 




We provide extensive experimental analysis and results on a  wide range of applications, combining a series of coarse and fine-grained material edits.
Results on both synthetic and real-world images demonstrate highly plausible material editing using \methodname{} for material transfer as well as fine-grained control.
We compare \methodname{} against other image/material editing approaches when a baseline is available, of which both our quantitative and qualitative analysis shows superiority in performance.
As we keep the based diffusion model intact, we find that the learned editing directions using the shader-based synthetic dataset can generalize to various image styles, including anime and paintings.
Overall, \methodname{} has several favorable properties for material editing in images:
\begin{itemize}
    \item \textbf{Wide Range of Novel Editing Controls.} To the best of our knowledge, \methodname{} is the first approach to offer parametric control, exemplar-based guidance, and blending of materials all within one general framework.
    \item \textbf{Operates only in CLIP-Space.} The minimal tuning nature of our approach brings maximal flexibility in model selection and performing multiple edits in one go.
    \item \textbf{Robustness in Various Styles.} \methodname{} can not only generate and edit materials of realistic images but can also be incorporated with various painting and artwork styles, bringing much flexibility to graphic designers.
\end{itemize}


\section{Related Work}

\textbf{Controlled Image Editing with Diffusion Models.} Recent advances in text and class-conditional image generation using diffusion models enable photorealistic image generation~\cite{dhariwal2021diffusion,ho2020denoising,ho2022cascaded,ho2022classifier,song2019generative,karras2022elucidating,kang2023scaling,saharia2022photorealistic,ramesh2022hierarchical,rombach2021high,nichol2021glide}. These models act as a base model for performing 3D-aware inpainting~\cite{rombach2021high,pandey2024diffusionhandles}, text-based editing~\cite{brooks2023instructpix2pix,ge2023expressive}, and controlled generation~\cite{ruiz2022dreambooth,kumari2023multi,chen2023subject}. High-level semantic and stylistic edits are based on inputs such as text-based instructions~\cite{hertz2022prompt,voynov2023p+,ge2023expressive,cao2023masactrl}, semantic segmentation~\cite{bar2023multidiffusion}, bounding box~\cite{li2023gligen,chen2024training,yang2023reco,wang2024instancediffusion}, and images~\cite{ruiz2022dreambooth,shah2025ziplora,ye2023ip,chen2024subject,rout2024rb,wang2024instantstyle}. 
InstructPix2Pix allows for instruction-based editing of images, allowing for stylistic and high-level semantic changes in the images~\cite{brooks2023instructpix2pix}. These methods are trained and hence limited to domains of high-level semantic changes, failing to edit low-level details such as object geometry and materials.
Beyond high-level semantic control, edits can be performed based on mid-level features such as depth maps~\cite{zhao2024uni,bhat2023loosecontrol} and edge-maps~\cite{zhang2023adding,mou2023t2i}.

\begin{figure}[t]
    \centering
    \includegraphics[width=\linewidth]{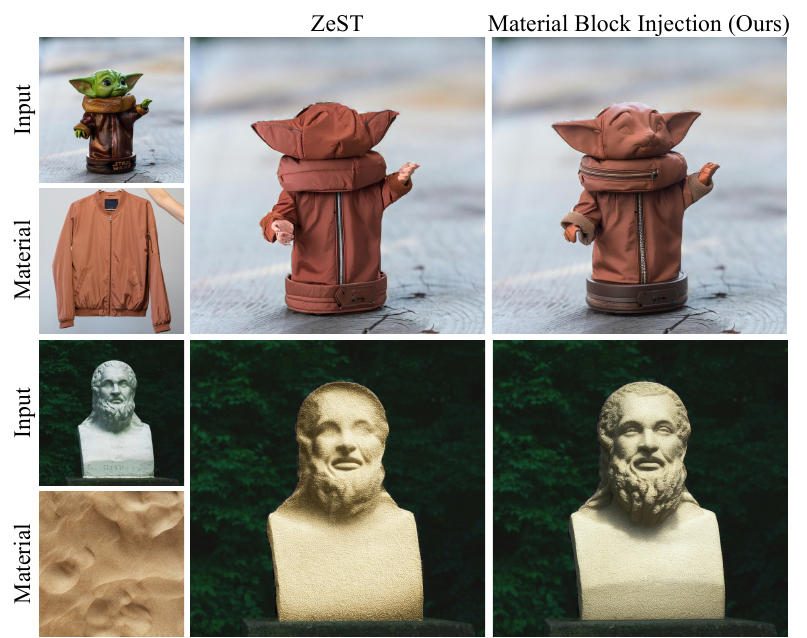}
    \caption{\textbf{Comparison of material block injection vs. all Blocks injection.} We present examples of using the same input and material exemplar. Given the same depth condition, injecting only into the material block allows much better geometry preservation compared to injecting to all blocks in the UNet. 
    }
    \vspace{-1em}
    \label{fig:comparison_style_zest}
\end{figure}

Recent work has enabled fine-grained continuous control enhancing the level on control in image editing~\cite{parihar2025precisecontrol,baumann2024continuous,gandikota2023concept}. Continuous control over concepts such as weather, age, and styles can be achieved in diffusion model-based generative models from a small set of text or images~\cite{gandikota2023concept}. Baumann et al. identifies directions within token-level CLIP text embeddings allowing for fine-grained over high-level attributes such as age and aesthetics in text-to-image models~\cite{baumann2024continuous}. These methods demonstrate control over high-level semantics using embeddings of pre-trained models.

\textbf{Material Editing.} Material editing is a challenging task, requiring understanding of object and scene properties such as geometry, illumination, and material attributes. Material acquisition methods extract material properties under known illumination and camera configurations~\cite{aittala2013practical,aittala2015two,deschaintre2019flexible}. Recent methods explore material recognition and segmentation with a data driven approach requiring little to no prior knowledge about the environment~\cite{bell2015material,liang2022multimodal,upchurch2022dense,sharma2023materialistic,kocsis2024intrinsic}.

Khan et al. proposed in-image material editing with normal estimates as approximations of the scene geometry~\cite{khan2006image}. Advances in generative models have facilitated more robust and photorealistic material editing techniques in images and 3D models~\cite{subias2023wild,delanoy2022generative,sharma2023alchemist,cheng2025zest,yeh2024texturedreamer,chen2023text2tex,richardson2023texture,guerrero2024texsliders,cao2023texfusion,lopes2024material,ceylan2024matatlas}. Coarse material editing in a zero-shot manner leveraging generative priors of pre-trained text-to-image models along with geometric and illumination information~\cite{cheng2025zest}. Fine-grained material properties can be edited by finetuning a generative model on physically rendered data~\cite{sharma2023alchemist}.

In our work, we propose a method for using using CLIP-space for material editing, specifically blending materials and recomposing fine-grained material attributes. 

\section{Method}


Our method, \methodname{}, uses CLIP embeddings and a pre-trained diffusion model to perform efficient material transfer, material blending, and parametric tuning of fine-grained material attributes. Specifically, we extend the architecture from ZeST, a zero-shot approach on performing exemplar-based material transfer by Cheng et al.~\cite{cheng2025zest}. 

Exemplar-based material transfer methods aim to transfer the material $M$ from a given exemplar image $I_m$ to an object in an input image $I$. ZeST performs the material transfer in a zero-shot manner using a pre-trained inpainting model (e.g., Stable Diffusion XL~\cite{rombach2021high}) $\mathcal{S}$. It guides the inpainting model using the foreground mask of the object $F_I$, depth map $D_I$ as geometric cue, and a foreground grayscale initial image $I_{init}$ as illumination cue, aiming to utilize only the material features $f(z_m)$ from $I_m$ during the generation process.

\begin{equation}
    I_{gen} = \mathcal{S}(I_{init}, F_I, D_I, f(z_m))
\end{equation}

Note that $f(\cdot)$ is the cross attention injection of feature $z_m$ originally computed using the CLS token from CLIP encoder with a fine-tuned head provided by IP-Adapter.
While this method results in images with plausible material transfer, this approach is not robust due to the convoluted nature of the CLIP embeddings -- some information besides materials is still passed to the denoising process, leading to cases of shifts in object geometry and shading. 

\begin{figure*}[]
    \centering
    \includegraphics[width=\linewidth]{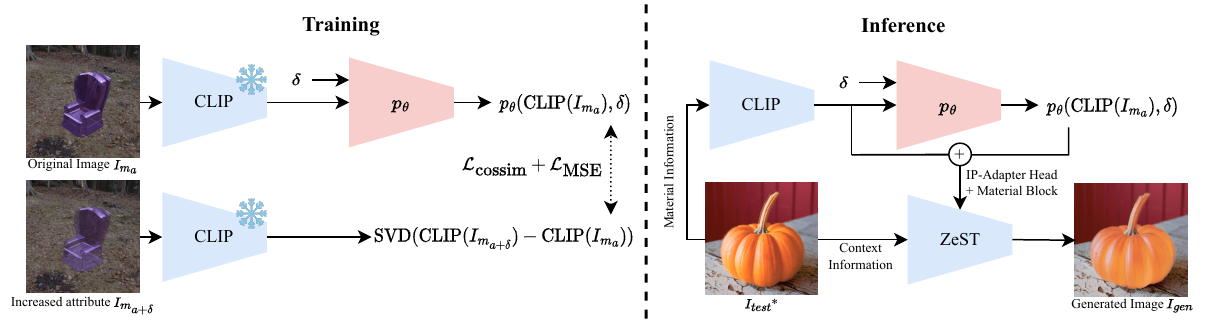}
    \caption{\textbf{Method overview for parametric material attribute control.} During training, we aim to learn $p_\theta$, a shallow MLP that predicts the editing direction in CLIP space given an image $I_{m_a}$. During inference, we can use $p_\theta$ to predict the offset that can be added to the CLIP embedding for parametric control. Note that $I_{test}$* during test time can be separated into two images, one for the context information (background, shading, geometry) and another for material.
    }
    \vspace{-1em}
    \label{fig:method}
\end{figure*}
\begin{figure}[t!]
    \centering
    \includegraphics[width=\linewidth]{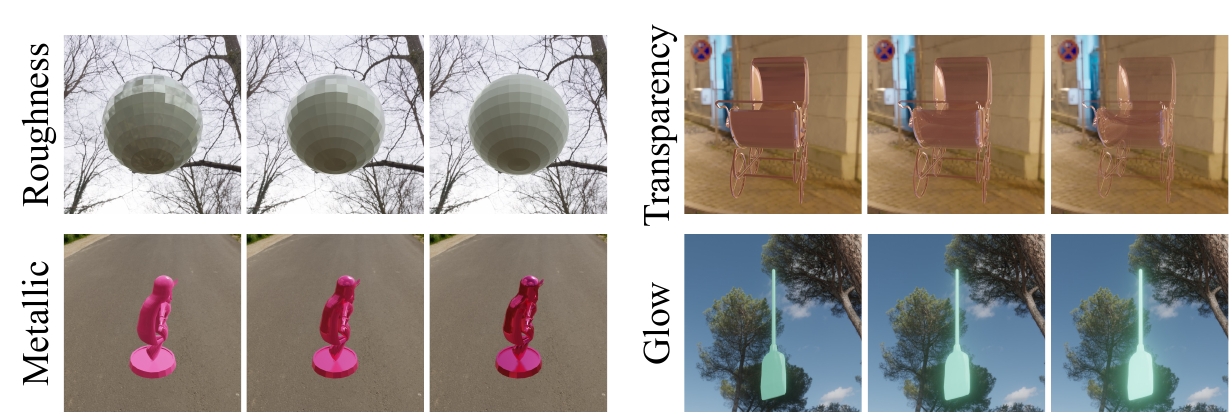}

    \caption{\textbf{Examples of Dataset.} We show samples from the rendered dataset for varying roughness, metallic, transparency, and glow.}
    \vspace{-1em}
    \label{fig:dataset}
\end{figure}

\subsection{Targeted Material Block Injection}
To mitigate these limitations, we find and inject the material embedding only to attention blocks in the denoising UNet of the inpainting model, the one responsible for attributing materials on the objects. Instead of injecting the material embedding $z_m$ at each of the attention layers in the denoising UNet, we find specific block responsible for material attribution following the process inspired by InstantStyle~\cite{wang2024instantstyle}. InstantStyle identifies a specific block responsible for injecting style information to the objects. We perform a similar study of exhaustively visualizing the generated results when injecting the information across each block of the denoising UNet to identify which layer contributes specifically to material transfer.
Our results illustrate that both material and style attribution on objects is performed by the same layer close to the bottleneck of the UNet. This is an expected outcome as material transfer can be seen as a specialized form of style transfer.


To this end, we propose to alter $f(\cdot)$ to inject the material embedding $z_m$ only in that specific block of denoising UNet. Figure \ref{fig:comparison_style_zest} presents examples comparing material block injection against all blocks injection (proposed by ZeST). In all examples, the geometry of the initial input condition is better preserved when the features are injected only into the material block. Note in Row 1 that the material transfer preserves the details of the material exemplar, while the original result of ZeST hallucinates hands on the toy figure -- a result primarily caused by the entanglement of the identity of the jacket and material.
This modification helps in preserving the geometry and lighting of the object and thus acts as the base architecture for \methodname{}.

\subsection{Material Blending}

Using this improved architecture, we aim to edit the context image with a material interpolated between two material exemplars. We observe that interpolating features from two material exemplars is also interpretable within the CLIP embeddings, similar to many results on finding interpretable directions in pre-trained models for pose and appearance~\cite{harkonen2020ganspace,pan2023drag}. This enables blended materials for image editing given two material features $z_{m_1}$ and $z_{m_2}$ extracted from two images using the CLIP encoder:
\begin{equation}
    I_{gen} = \mathcal{S}(I_{init}, F_I, D_I, f(\alpha z_{m_1} + (1 - \alpha) z_{m_2})),
\end{equation}
where $\alpha > 0$ is the interpolation weights.

Material blending can be performed with three different configurations of the exemplar images: (1) different objects and materials, (2) different objects made of the same materials with a single attribute (e.g. roughness) varied, and (3) same object, same materials with a single attribute varied.


\subsection{Parametric Control from a Single Image}
In addition to blending between two materials from two exemplar images, we explore the use of CLIP-space embeddings for achieving parametric control over fine-grained material attributes. Specifically, we demonstrate parametric control over roughness, metallic, transparency, and glow. 

Given a material exemplar image $I_{m_a}$ with a specific material attribute $a$, and an editing strength $\delta$, we train a attribute editing network $p_\theta$ to predict the corresponding CLIP feature of $I_{m_{a+\delta}}$. We train the attribute editing network for each attribute individually using a synthetically rendered dataset.
Next, we describe the dataset preparation, training, and inference setup of our method.

\vspace{1mm}
\noindent\textbf{Dataset Creation.}
Since collecting real world dataset with controlled material attribute changes is impractical, we render a small  dataset using Blender with controlled shader properties. Contrary to the approach of Alchemist~\cite{sharma2023alchemist}, our model only predicts directions in the CLIP-space and thus requires much less data. We show in Section \ref{sec:num_objs} an ablation on the quality of generated images against the number of objects used, where our attribute changing network could be learned with even as few as 8 objects.

We used 300 synthetic objects~\cite{deitke2023objaverse} (250 for training and 50 for validation) and pair each with a random HDR map from a collection of 50 maps. To create a dataset for attribute $a$, we create a default material per object with attributes other than $a$ randomly assigned. Then, we render the object at a random viewpoint with traversing the value of $a$ from uniform steps. Note that for transparency, we not only increase the transmission value of the material but also decrease the roughness effect to create a more glass-like transparent appearance. We present some examples in Figure \ref{fig:dataset}.

\vspace{1mm}
\noindent\textbf{Training and Inference Setup.}
While this rendered dataset proved to be useful for our task, we identify two key limitations. First, the dataset is fairly small resulting in potential inductive biases in the image features. Second, note that CLIP features are fairly noisy~\cite{lan2024clearclip}. These observations suggest that there may be a small set of features within the CLIP features we should not learn from. To mitigate this, we stack the editing directions of a given attribute (computed as the difference of two CLIP features given an image pair) and perform singular value decomposition to obtain a low-rank approximation of the stacked matrix. The rank for each attribute is decided by the elbow method when plotting out the singular values. The variance explained for all four attributes fall within the range of $67\% - 80\%$. 


Figure \ref{fig:method} provides an overview of the training and inference setup. During training, we take an input image $I_{m_a}$ and an editing strength $\delta$. We then train our attribute editing network $p_\theta(I_{m_a}, \delta)$ (a 2-layer MLP) with the criterion:
\begin{multline}
    \arg \min_\theta[\mbox{cossim}(s_{m_{a+\delta}} , p_\theta(I_{m_a}, \delta)) \\ 
    + \mbox{MSE}(s_{m_{a+\delta}}, p_\theta(I_{m_a}, \delta))],
\end{multline}
where $s_{m_{a+\delta}}$ is the low-rank material attribute from $a$ to $a+\delta$ approximated by SVD, $[\mbox{cossim}, \mbox{MSE}]$ are the cosine similarity loss and mean-squared loss, respectively. With this objective, $p_\theta$ learns to predict the low-rank approximated CLIP feature of the same original image with one attribute $\alpha$ increased by $\delta$.


With a learned attribute editing network $p_\theta$ at inference time, we can obtain fine-grained material features after tuning attribute $a$ to $a+\delta$ altering $z_{m_a}$ into $z_{m_{a+\delta}}$:
\begin{equation}
    z_{m_{a+\delta}} = \mbox{CLIP}(I^i_m) + p_\theta(I^i_m, \delta). 
\end{equation}

This feature allows us to build on top of our exemplar-based transfer pipeline, where we can regenerate an image with recomposed material attributes. 

Since we did not finetune the pre-trained diffusion model, each attribute network can be trained separately and used jointly to find the designated CLIP feature. We provide examples of this in Section \ref{sec:discussion}.





\section{Experiments}
We present qualitative and quantitative evaluations to validate \methodname{}. We analyze the effectiveness of our method at material blending and fine-grained parametric control over material attributes. Further analysis on the robustness beyond natural images and dataset efficiency for achieving parametric control demonstrates the practical impact for material editing. 
\begin{figure}[]
    \centering
    \includegraphics[width=\linewidth]{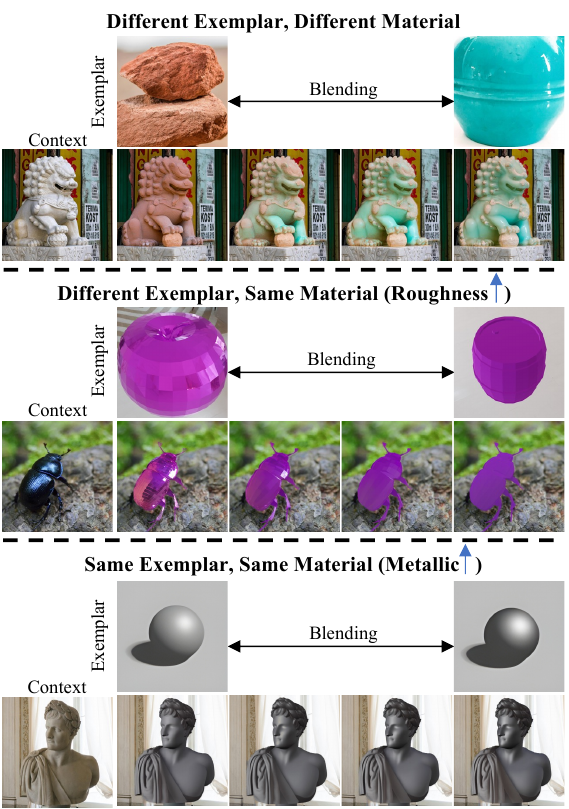}
    \caption{\textbf{Material blending results.} By interpolating the CLIP features of the material exemplars, \methodname{} can transfer the intermediate blended features to the input image, creating material blending effect. Blending can work with exemplar images with following configurations: (1) Different objects with different materials, (2) Different objects made of the same base material except one varying attribute (metallic), and (3) Same object and same material with one attribute (metallic) varying.}
    \vspace{-1em}
    \label{fig:interpolation}
\end{figure}

\begin{figure*}[]
    \centering
    \includegraphics[width=\linewidth]{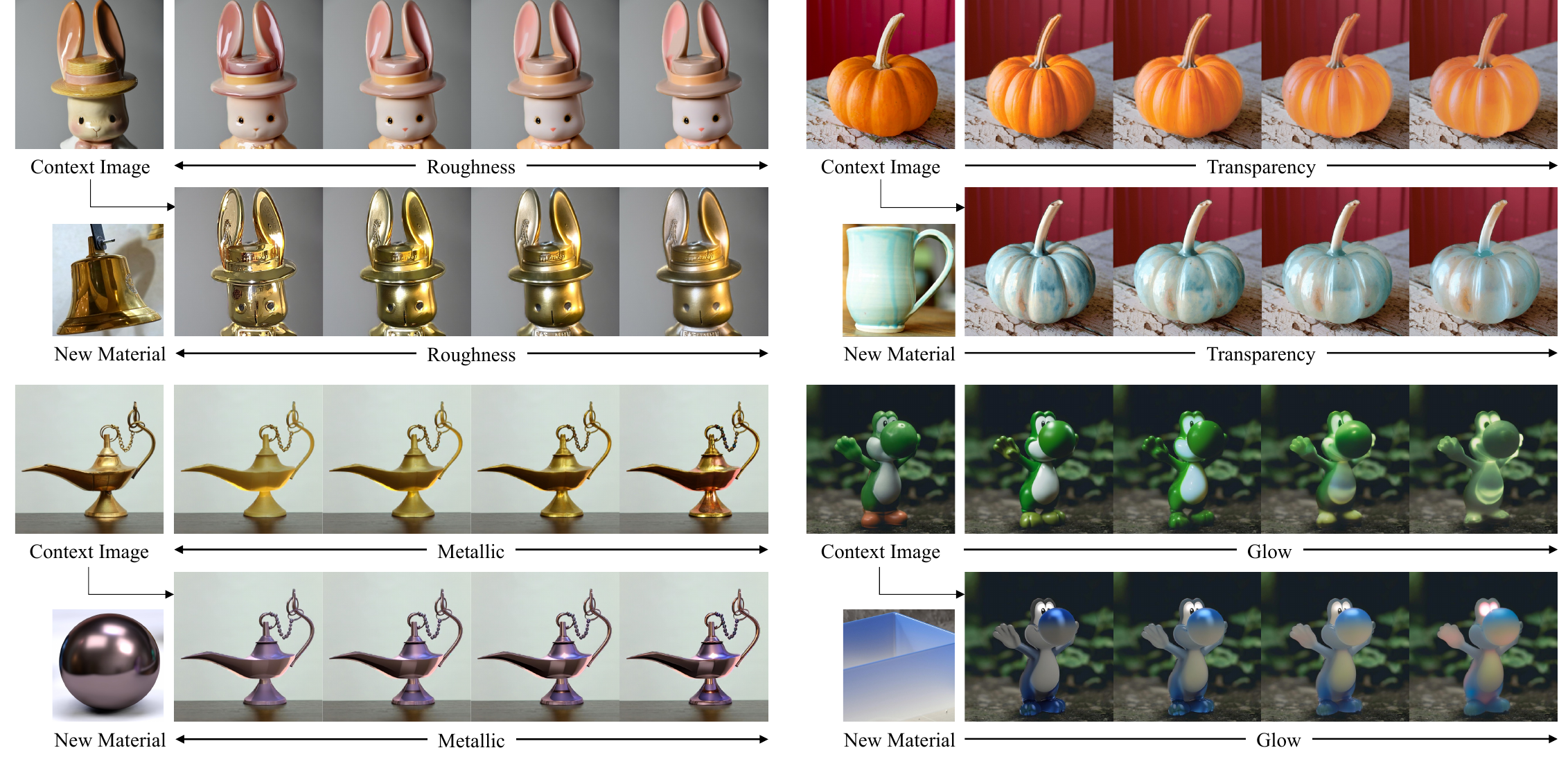}
    \vspace{-2em}
    \caption{\textbf{Parametric control results.} We present four sets of results controlling roughness, transparency, metallic, and glow. For each set of results, we present one example directly using the reference image for context and material, and another set where we change to a new material exemplar. \methodname{} disentangles the reflections from the albedo to provide perceptually convincing results.}
    \vspace{-1em}
    \label{fig:slider_results}
\end{figure*}

\subsection{Qualitative Results}
\noindent\textbf{Material Blending.}
We present material blending results with different selection of material pairs in Figure \ref{fig:interpolation}. Note that exemplar materials $m_1$ and $m_2$ can be in different configurations. They can be two completely different material exemplars (example 1) or the same exemplar and material with only one attribute varying (example 3). Surprisingly, even when the two exemplars are of different objects with the same base material with a single varying attribute (second example), the CLIP embeddings are sufficiently able to identify the underlying attribute and perform parametric control through material blending. 

\vspace{1mm}
\noindent\textbf{Parametric Control of Material Attributes.}
We present slider results for roughness, metallic, transparency, and glow in Figure \ref{fig:slider_results}. For each of the attributes, we show two sliding examples, one using the reference image for both context and material, and the other with a new material applied demonstrating the combination of material editing and fine-grained control in a single forward pass. 

Note that the attribute we intend to control is disentangled from the other attributes. For the roughness examples, we observe reduction in the specularity on the surface as roughness increases. For the transparency and metallic examples, the reflection is disentangled with the albedo/base color of the object, lighting up the colors in some regions and darkening the others. For the glow example, the color of the glow follows the original albedo of the object.


\noindent\textbf{Parametric Control Qualitative Comparisons.}
We compare our method to three baselines, namely InstructPix2Pix~\cite{brooks2023instructpix2pix}, Concept Slider (Text), and Concept Slider (Image)~\cite{gandikota2023concept}. 


For InstructPix2Pix, we use prompts to guide attribute changes. Specifically, given an image, we use the prompt \texttt{``Make the *object {more/less} *attribute''}, where \texttt{*object} is the object in the image and \texttt{*attribute} is the intended attribute change (e.g., \texttt{Make the chair transparent}). Note that this does not allow for parametric control.

We also implement two types of Concept Sliders using text and image with the SDXL backbone. For text concept sliders, we find opposite words describing the attribute (e.g. \texttt{transparent} and \texttt{opaque}, \texttt{rough} and \texttt{smooth}) and train a slider for each set of attributes. For image sliders, we use pairs from the two ends of the spectrum of our dataset to train for each material attribute. During inference, we perform DDIM inversion on the image and increase the slider value to the maximum before noticeable artifacts occur.

\begin{figure*}[]
    \centering
    \includegraphics[width=\linewidth]{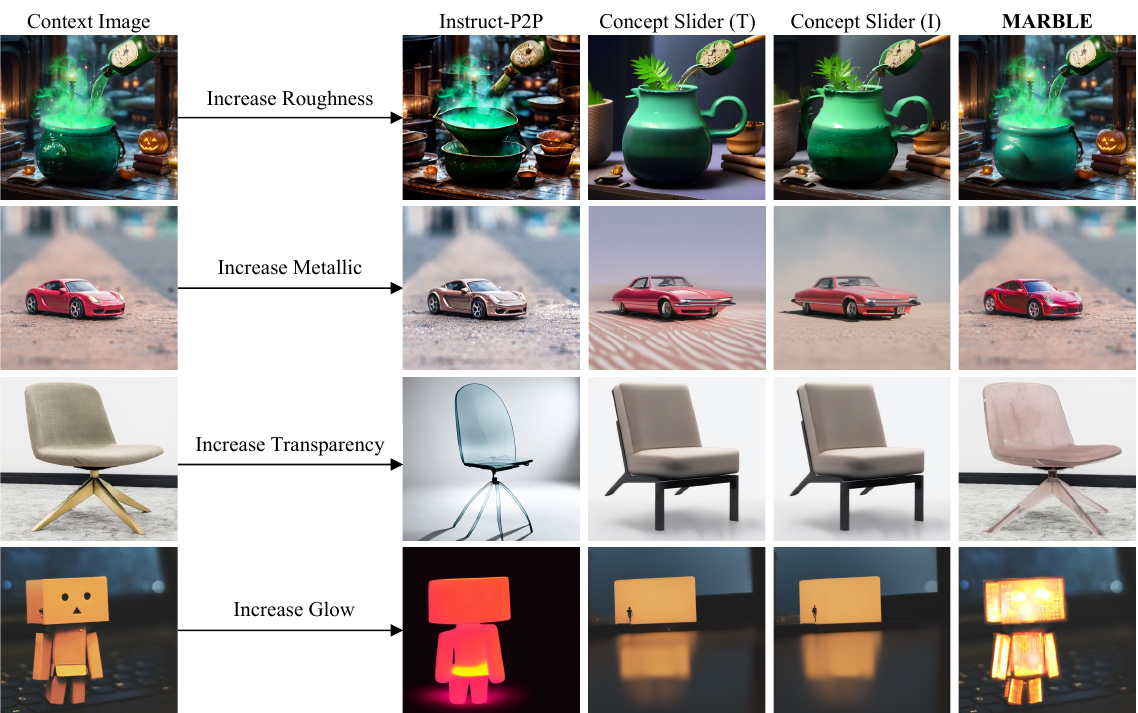}
    \caption{\textbf{Qualitative comparisons.} We compare against InstructPix2Pix and 2 versions of Concept Sliders. (T) and (I) denote text and image trained versions, respectively. All baselines either fail to capture the parametric control (Concept Sliders/chair/transparency), or result in undesired changes in object geometry (toy figure/glow, pot/roughness) or albedo (InstructPix2Pix/car/metallic).}
    \label{fig:comparison_results}
\end{figure*}

Figure \ref{fig:comparison_results} presents the qualitative comparisons against the baselines. Due to the ambiguity of text descriptions, InstructPix2Pix often leads to unintended changes on other attributes such as the geometry of the pot and chair, the albedo of the car for metallic, or the background for the toy figure for glow. On the other hand, editing with Concept Sliders (trained with either text or image pairs) requires DDIM inversion in the first place, which leads to inaccurate reconstructions even without parametric changes. While the sliders occasionally produces reasonable outputs for metallic and roughness (rougher pot surfaces and reflection on the car), the concept of transparency and glow were not captured by this approach. Our method produces high-fidelity results, showing disentangled and accurate edits for all attributes.

\subsection{Quantitative Results}

Out of the three baselines, only the image-trained concept slider allows us to compute quantitative metrics. InstructPix2Pix does not allow continuous control and neither does the editing strength of text-trained concept slider correspond well with the actual shader value changes in Blender.

Using a rendered validation set comprising of 50 objects, each with changing material attributes, we compare against image-trained concept sliders in terms of PSNR, LPIPS~\cite{zhang2018unreasonable}, CLIP Score~\cite{radford2021learning} and DreamSim~\cite{fu2023dreamsim}. Table \ref{tab:quantitative} shows that \methodname{} performs better than baselines across all metrics for all attributes.

\begin{table}[t!]
\vspace{-1em}
    \caption{\textbf{Quantitative comparisons.} We present quantitative comparisons for all attribute controls compared to Concept Sliders trained using our dataset.}
    \vspace{-1em}
    \centering
    \resizebox{\linewidth}{!}{

    \begin{tabular}{lcccc}
    \toprule
    & PSNR$\uparrow$ & LPIPS$\downarrow$ & CLIP$\uparrow$ & DreamSim$\downarrow$\\
    \midrule
    \rowcolor[gray]{0.9} \textbf{Roughness} &&&&\\
    Concept Slider (Images)  & 18.87& 0.356 & 0.597& 0.567\\
    \methodname{} & \textbf{26.56} & \textbf{0.056} & \textbf{0.931}& \textbf{0.129}\\  
    \midrule
    \rowcolor[gray]{0.9} \textbf{Metallic} &&&&\\
    Concept Slider (Images)  & 19.45&  0.317& 0.655& 0.479\\
    \methodname{} & \textbf{26.82} & \textbf{0.053} & \textbf{0.928}& \textbf{0.121}\\  
    \midrule
    \rowcolor[gray]{0.9} \textbf{Transparency} &&&&\\
    Concept Slider (Images)  & 19.85&  0.346& 0.639& 0.525\\
    \methodname{} &\textbf{26.99} & \textbf{0.070} & \textbf{0.905} & \textbf{0.163}\\  
    \midrule
    \rowcolor[gray]{0.9} \textbf{Glow} &&&&\\
    Concept Slider (Images)  & 16.92& 0.301& 0.661& 0.509\\
    \methodname{} &\textbf{ 19.73}& \textbf{0.111}& \textbf{0.890}&\textbf{0.213}\\  

    \bottomrule
    \end{tabular}
    }
    \vspace{-1em}
    \label{tab:quantitative}
\end{table}

\vspace{1mm}
\noindent\textbf{User Study.} To further validate the effectiveness of \methodname{} on real-world images, we conduct a user study with 16 participants. We generate results on 20 real-world images with edits controlling a random material attribute using our method and image-based Concept Slider. Each user was provided 3 image sets to compare based on the intended control. As a result, $87.5\%$ participants chose images generated by our method, \methodname{}.


\subsection{Discussion}
\label{sec:discussion}
\begin{figure}[]
    \centering
    \includegraphics[width=0.95\linewidth]{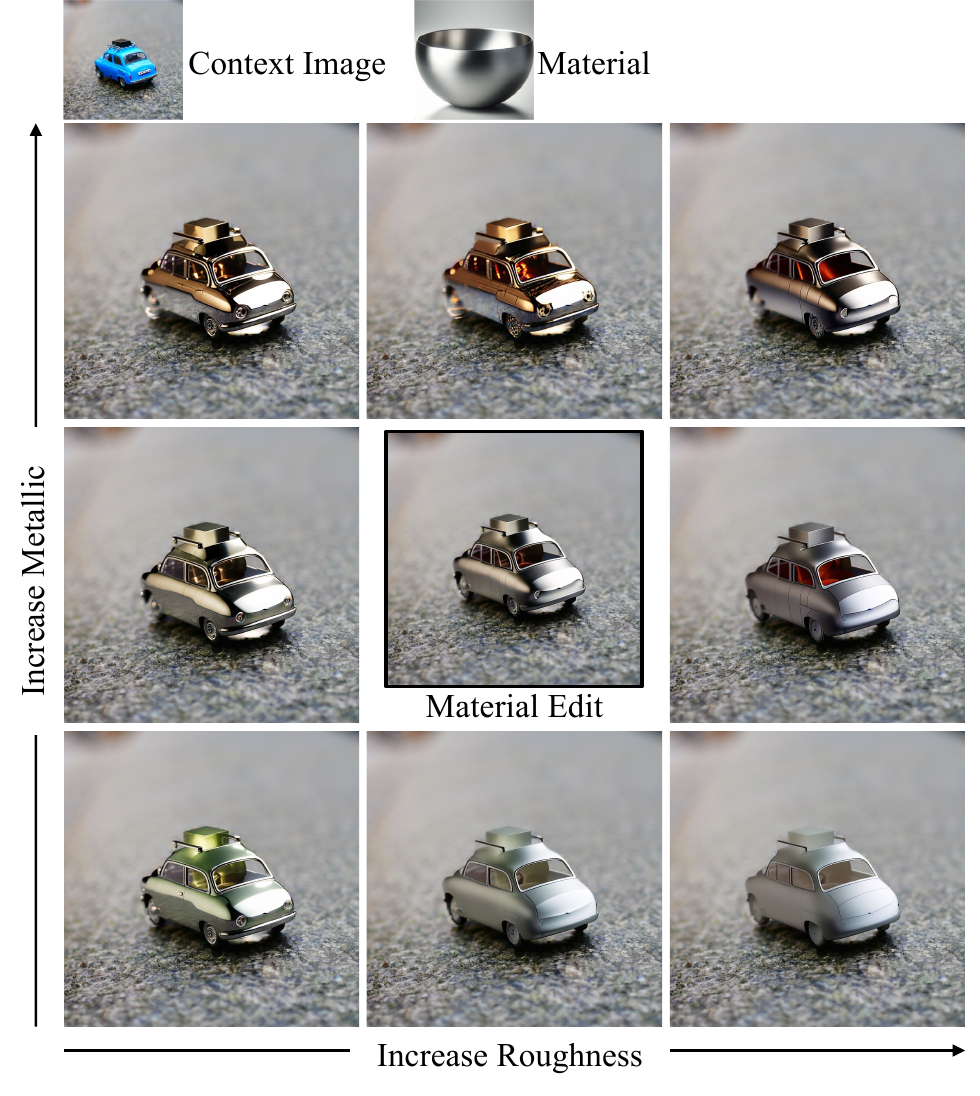}
    \vspace{-1em}
    \caption{\textbf{Multiple controls at once}. With minimal tuning on the pre-trained components, \methodname{} can perform material transfer and incorporate multiple attribute controls all in a single pass on real-world images. We present a grid of results of increasing roughness and metallic of a toy car, where we can see that the two attributes are properly disentangled from one another.}
    \vspace{-1em}
    \label{fig:multi_concept}
\end{figure}

\noindent\textbf{Multi-Concept Control-Grid.}
One of the main merits of CLIP-based control is the ability to control multiple attributes in a single forward pass. Figure \ref{fig:multi_concept} presents
an example of controlling roughness and metallic components of the toy car's material. \methodname{} allows us to transfer a metal material onto the toy, while simultaneously enabling fine-grained control over metallic and roughness of material. Each image is generated in a single forward pass. While trained separately, we can see that the two attributes are disentangled from one another even when applied together. 

\noindent\textbf{Robustness on Real-World Images.} 
\begin{figure}[]
    \centering
    \includegraphics[width=\linewidth]{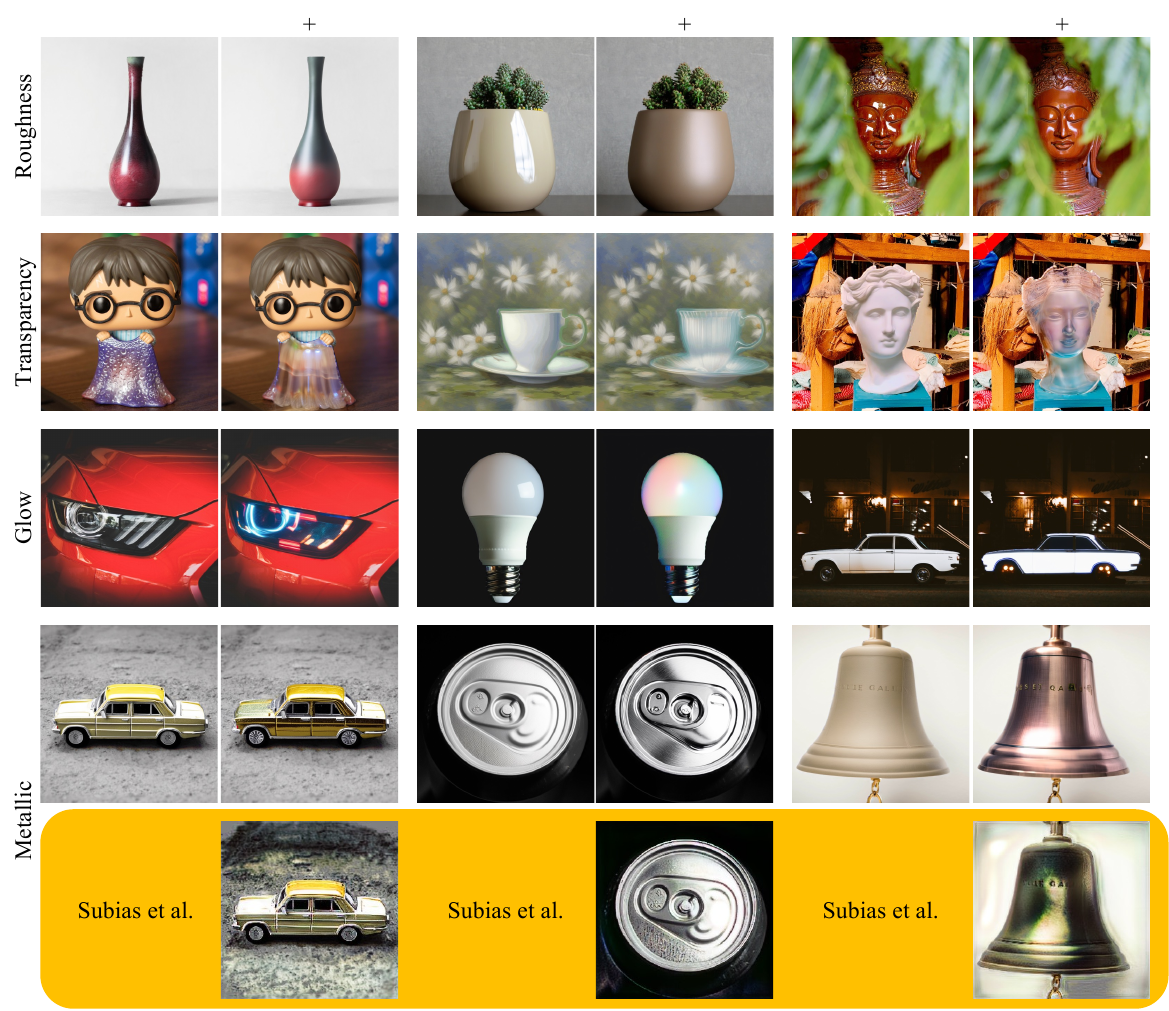}
    \caption{\textbf{Additional Results for Attribute Control.} We present 12 pairs of results on increasing attribute value (From left to right). As In-the-wild editing by Subias et al. also support metallic, we show the qualitative comparisons for the three examples. \textit{Zoom in for details.} 
    }
    \vspace{-2em}
    \label{fig:additional_results}
\end{figure}
In addition to the eight slider examples, we also present a variety of results of increasing the value one attribute. MARBLE was able to perform realistic edits across a variety of object from different backgrounds. As In-the-wild editing by Subias et al. also support metallic, we show the qualitative comparisons for the three examples given.

\vspace{1mm}
\noindent\textbf{Parametric Control with Different Styles.}
\begin{figure}[]
    \centering
    \includegraphics[width=\linewidth]{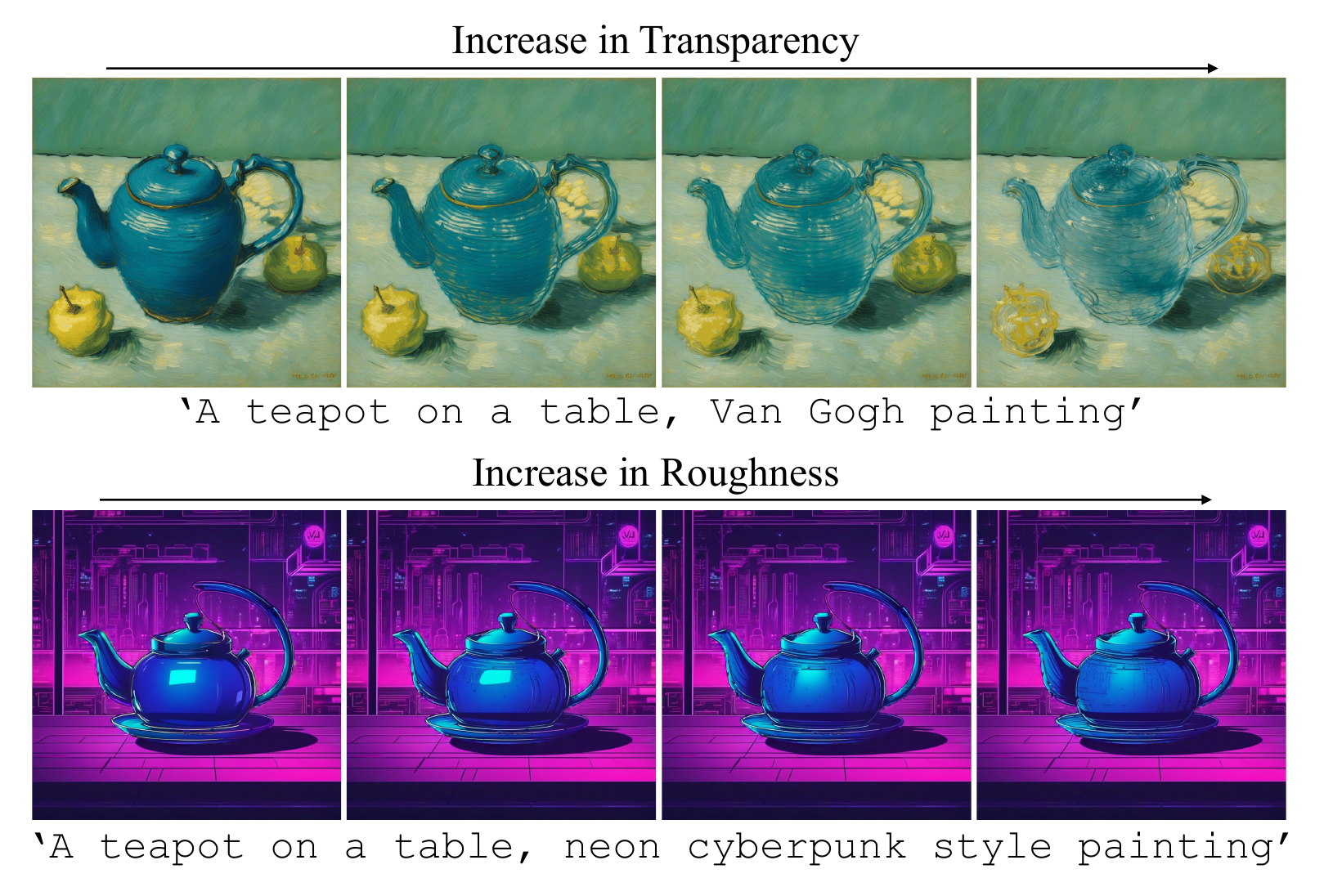}
    \vspace{-2em}
    \caption{\textbf{Parametric control with different styles.} By leveraging the generalization capability of CLIP, our parametric controls can be also be adopted for images with various styles. We present parametric control over two styles of paintings generated by SDXL. Despite being trained on rendered images, the parametric controlled editing preserves the given style when presenting attribute changes.}
    \vspace{-1em}
    \label{fig:styles_qualitative}
\end{figure}
By solely operating in the CLIP space and not changing the pre-trained weights of the base diffusion model, \methodname{} also shows capabilities to perform parametric control on various styles understood within the CLIP text features.
Figure \ref{fig:styles_qualitative} presents two examples of paintings with different styles generated by SDXL. \methodname{} changes the transparency and roughness of the foreground object while preserving the original style of the image. This is particularly evident on the wiggly brush strokes on the transparent pot mimicking the style of Van Gogh.


\vspace{1mm}
\noindent\textbf{How small can the training dataset be?}
\label{sec:num_objs}
\begin{figure}[]
    \centering
    \includegraphics[width=\linewidth]{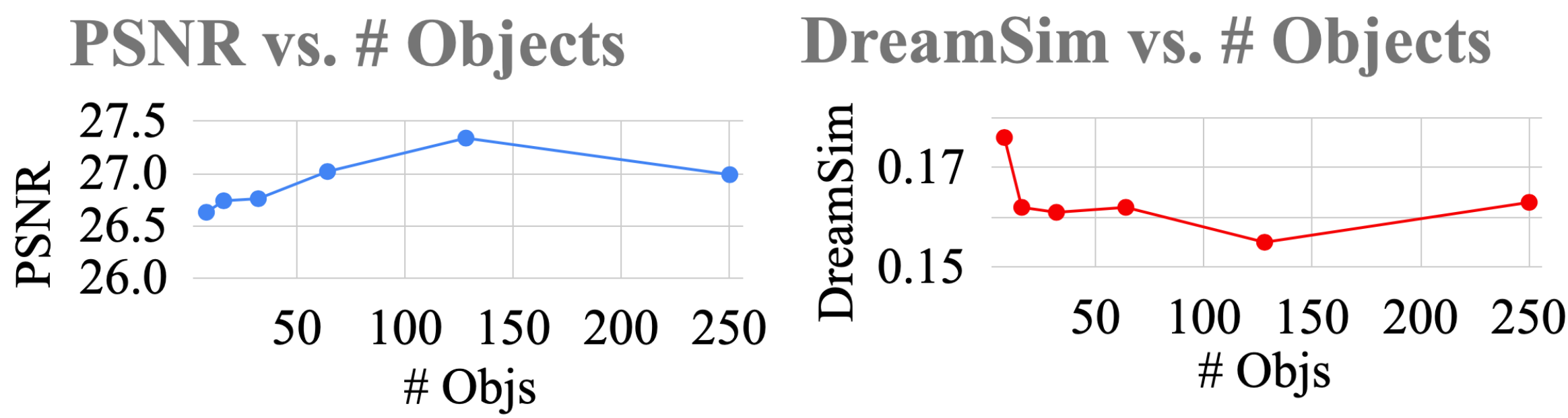}
    \caption{\textbf{Data efficiency.} We trained the transparency controller with 8, 16, 32, 64, 128 and 250 objects and present their PSNR and DreamSim scores. Even with as few as 16 objects, we can still obtain decent numbers on the validation dataset.}
    \label{fig:num_objs}
    \vspace{-1em}
\end{figure}
Furthermore, we investigate how small the training dataset can be by measuring the PSNR and DreamSim on synthetically rendered validation dataset (Figure \ref{fig:num_objs}). To our surprise, training on as few as 16 objects was sufficient to achieve similar results compared to using the full dataset. Qualitative results on real-world dataset are also presented in the Appendix.

\vspace{1mm}
\noindent\textbf{Limitations.}
\begin{figure}[]
    \centering
    \includegraphics[width=\linewidth]{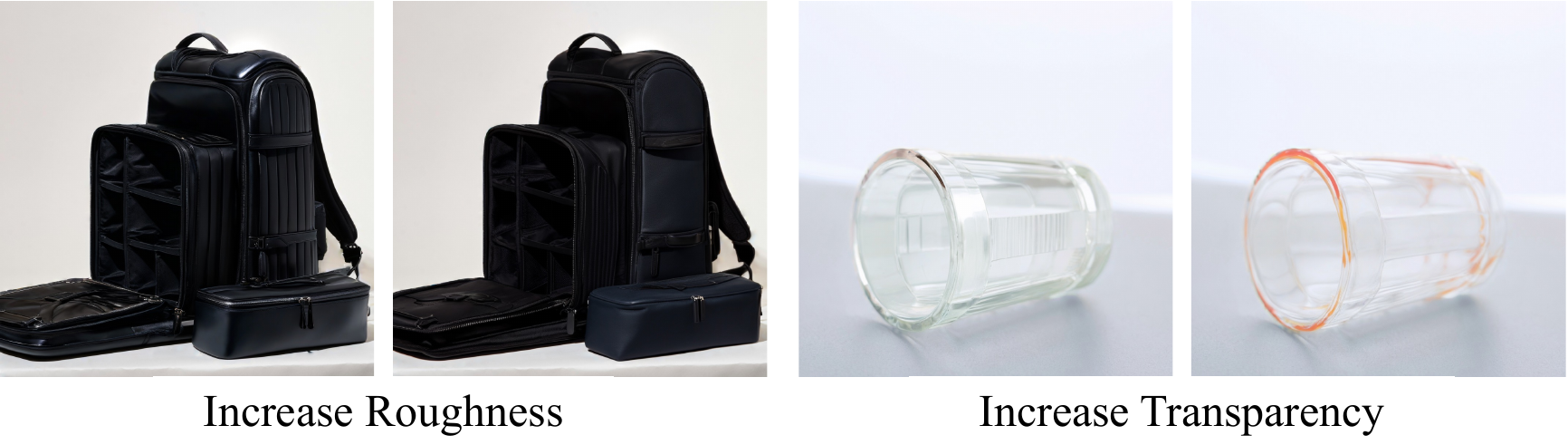}
    \vspace{-1em}
    \caption{\textbf{Limitations.} Our method has two primary limitations. (1) Sometimes performing parametric control also changes the texture patterns of the object such as the pattern on side of the leather backpack changes as roughness increases (left). (2) Sometimes the effects of the parametric control leads to artifacts.}
    \vspace{-2em}
    \label{fig:limitations}
\end{figure}
Our method has two main limitations, as shown in Figure \ref{fig:limitations}. First, parametric control would sometimes change the textural patterns of an object, such as the pattern on the leather backpack of the left example. Second, the effect of the control causes undesired artifacts when the model is not expected to result in no change, as observed in the case of increasing transparency of the glass. These artifacts and loss of the high-frequency details can be caused due to multiple reasons such as the effect of noise pattern added to the latent of the context image, operations in the noisy CLIP-space, or the information loss in the encoding-decoding process of SDXL.

\section{Conclusion}
We present \methodname{}, a method using CLIP-space for material editing in images. \methodname{} builds on top of previous works in parametric and exemplar-based control, while adding a new blending mechanism, to allow flexibility of users to blend and recompose materials in a given image. The controls can be trained without finetuning generative model, and allows for multiple edits in one single forward pass. Overall, \methodname{} presents an interesting direction for fine-grained, graphics-based controls of generative models revealing the advantages of CLIP-space representation for low-level controlled editing.
{
    \small
    \bibliographystyle{ieeenat_fullname}
    \bibliography{main}
}


\end{document}